\documentclass[runningheads]{llncs}

\usepackage[T1]{fontenc}
\usepackage{graphicx}
\usepackage{tikz,xcolor,hyperref}

\usepackage{makecell}
\usepackage{graphicx}
\usepackage{amsmath}
\usepackage{amssymb}
\usepackage{booktabs}
\usepackage{subfigure}
\usepackage{bm}

\usepackage{multicol, multirow}
\usepackage{dsfont}
\usepackage{xcolor}
\usepackage{orcidlink}

\begin{document}
%
% \title{Dynamic Spatial Feature Capture for Small Object Detection in Remote Sensing Images Using MKSNet}
\title{MKSNet: Advanced Small Object Detection in Remote Sensing Imagery with Multi-Kernel and Dual Attention Mechanisms}

\titlerunning{MKSNet-Multi-Kernel Selection Network} 
\author{Jiahao Zhang\inst{1,2} \textsuperscript{\orcidlink{0009-0009-0847-5805}} \and
Xiao Zhao\inst{3} \and
Guangyu Gao\thanks{Corresponding author, guangyugao@bit.edu.cn.}\inst{1,2}\textsuperscript{\orcidlink{0000-0002-0083-3016}}}
\institute{School of Computer Science, Beijing Institute of Technology, Beijing, China\\
\and
Tangshan Research Institute, Beijing Institute of Technology, Tangshan, China\\
\and
Shanghai Institute of Satellite Engineering, Shanghai, China
}
\maketitle% typeset the header of the contribution
\begin{abstract}
Deep convolutional neural networks (DCNNs) have substantially advanced object detection capabilities, particularly in remote sensing imagery. 
However, challenges persist, especially in detecting small objects where the high resolution of these images and the small size of target objects often result in a loss of critical information in the deeper layers of conventional CNNs. 
Additionally, the extensive spatial redundancy and intricate background details typical in remote-sensing images tend to obscure these small targets.
To address these challenges, we introduce Multi-Kernel Selection Network (MKSNet), a novel network architecture featuring a novel Multi-Kernel Selection mechanism.
The MKS mechanism utilizes large convolutional kernels to capture an extensive range of contextual information effectively. 
This innovative design allows for adaptive kernel size selection, significantly enhancing the network's ability to dynamically process and emphasize crucial spatial details for small object detection.
Furthermore, MKSNet also incorporates a dual attention mechanism, merging spatial and channel attention modules. 
The spatial attention module adaptively fine-tunes the spatial weights of feature maps, focusing more intensively on relevant regions while mitigating background noise. 
Simultaneously, the channel attention module optimizes channel information selection, improving feature representation and detection accuracy.
Empirical evaluations on the DOTA-v1.0 and HRSC2016 benchmark demonstrate that MKSNet substantially surpasses existing state-of-the-art models in detecting small objects in remote sensing images. 
These results highlight MKSNet's superior ability to manage the complexities associated with multi-scale and high-resolution image data, confirming its effectiveness and innovation in remote sensing object detection.
\end{abstract}


\section{Introduction}

In remote sensing image applications, accurately identifying small targets is crucial for monitoring terrestrial features, assessing environmental changes, and evaluating disaster impacts \cite{liu2016geological}.
However, detecting small targets within remote sensing imagery presents considerable challenges. 
Traditional backbone networks typically employ small convolutional kernels, such as ResNet \cite{he2016deep}, which fail to extract sufficient contextual information necessary for recognizing multi-scale targets.
Moreover, small targets often blend into their backgrounds due to similar textures and colors, complicating their distinction from background noise. 
Therefore, addressing these challenges is essential to enhance the accuracy and efficiency of detection methods.

To mitigate these issues, various strategies have been explored. 
Feature Pyramid Networks (FPN)~\cite{lin2017feature} and Atrous Spatial Pyramid Pooling (ASPP) techniques~\cite{wang2018dense} aim to enhance multi-scale target detection by integrating features across different scales. 
However, these methods often fall short when detecting very small targets due to the lack of detail in fused multi-scale features.
Other methods, such as background subtraction and saliency detection, increase the visibility of small targets but also raise computational demands and risk inaccuracies in complex environments.
Depthwise Separable Convolutions~\cite{chollet2017xception} (e.g., RetinaNet~\cite{lin2017focal}), while reducing computational complexity, may strip away crucial contextual details, thereby impairing the network's capability to capture fine details and background nuances of small targets. 
Although these techniques offer advantages under specific conditions, their limitations significantly impact overall performance in detecting small targets.

This paper proposes the Multi-Kernel Selection Network~(MKSNet), a novel network architecture specifically designed to address these limitations.
Our approach leverages large convolutional kernels to extract more comprehensive contextual information, crucial for the detection of small targets. 
By integrating convolutional kernels of various sizes, MKSNet significantly enhances its capability to manage multi-scale targets.
Furthermore, we introduce a dual attention mechanism that combines spatial and channel attention.
The spatial attention mechanism adaptively adjusts feature map weights to intensify focus on relevant regions while minimizing background noise.
Concurrently, the channel attention mechanism refines the weighting of feature channels to optimize feature representation and emphasize crucial attributes. 
This dual-attention framework not only maintains low complexity and high efficiency but also effectively counters the drawbacks of previous methods when dealing with complex backgrounds and high-resolution images, thus offering substantial benefits for remote sensing image processing.

Our main contributions are summarized as follows:
\begin{itemize}
    \item \textbf{Application of Large Convolutional Kernels:} We introduce the use of large convolutional kernels to extract more comprehensive contextual features. 
    This enhancement is particularly beneficial for small target detection, as large kernels capture richer contextual information, thereby improving the network's ability to detect small targets.

    \item \textbf{Multi-Scale Convolutional Kernels:} By employing convolutional kernels of varying sizes, we significantly enhance the network's performance in handling multi-scale targets. 
    This multi-scale kernel design facilitates more effective capture and fusion of features at different scales, thereby increasing detection accuracy.

    \item \textbf{Dual Attention Mechanism:} We incorporate both spatial and channel attention mechanisms. 
    The spatial attention mechanism adaptively adjusts the spatial weights of feature maps, thereby focusing more on critical regions and reducing background noise interference. 
    The channel attention mechanism optimizes the weights of feature channels to improve feature extraction. 
    This dual-attention approach not only maintains low complexity and high efficiency but also effectively addresses the limitations of existing methods in complex backgrounds and high-resolution images, demonstrating significant advantages in remote sensing image processing.
    
    \item \textbf{Enhanced Detection Performance:} Combining large convolutional kernels with the dual-attention mechanism, we effectively overcome the limitations of current methods. 
    Our innovations lead to substantial improvements in small target detection under high-resolution and complex background conditions, showcasing a notable advancement over traditional approaches.
\end{itemize}

\begin{figure}
    \centering
    \includegraphics[width=0.8\linewidth]{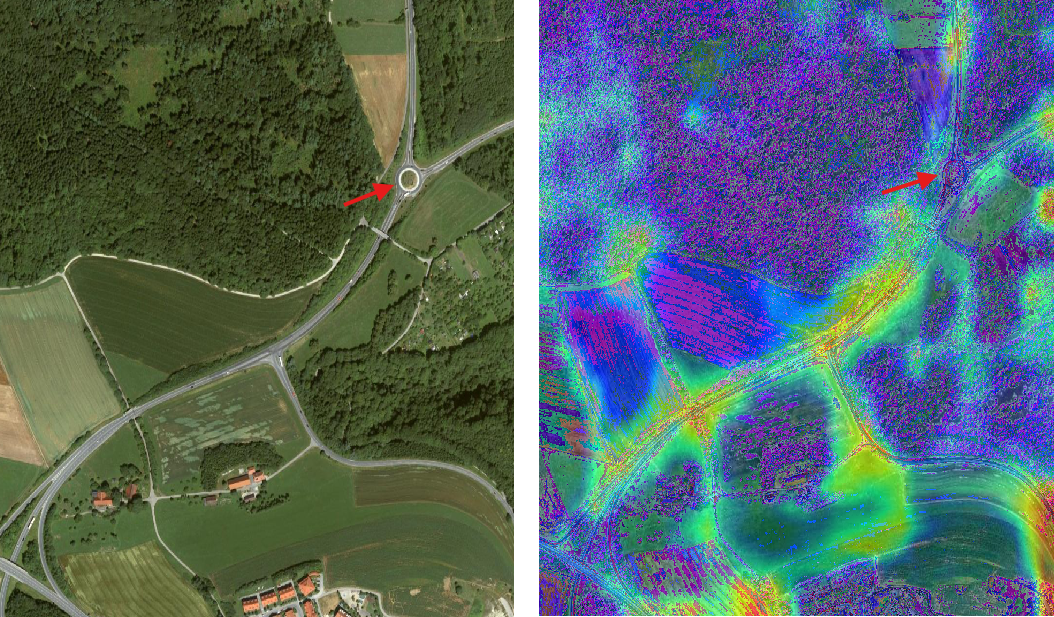}
    \caption{Roundabout Detection in DOTA-v1.0. 
    On the left, traditional networks struggle to distinguish the roundabout from similar structures such as intersections and storage tanks, due to small convolutional kernels. 
    The right image shows a heatmap from MKSNet, highlighting its superior ability to capture contextual details and accurately identify the roundabout.}
    \label{fig:enter-label}
\end{figure}

\begin{figure}
    \centering
    \includegraphics[width=0.8 \linewidth]{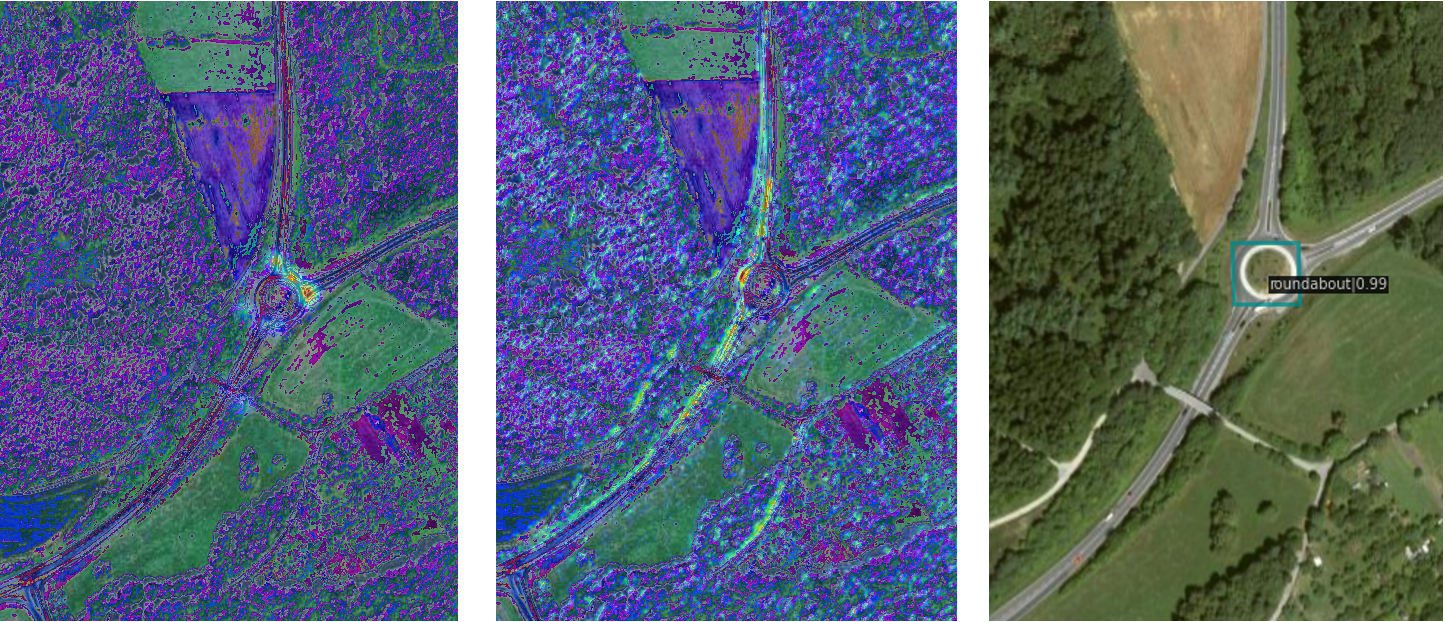}
    \caption{Comparative Heatmaps and Detection Results for Roundabout Recognition. 
    The first heatmap shows a network with fewer large kernels focusing on local features of the roundabout, while the second highlights a network with more large kernels capturing broader contextual information. 
    The third image demonstrates MKSNet's accurate and rapid recognition of the roundabout, leveraging enhanced contextual understanding.}
    \label{fig:enter-label}
\end{figure}

\section{Related Work}

\subsection{Large Convolutional Kernels}

Large convolutional kernels have recently gained prominence.
These kernels are particularly effective at capturing broad contextual information, which is crucial for the detection of small objects.
For example, RepLKNet~\cite{ding2022scaling} modifies the architecture of Swin Transformer~\cite{liu2021Swin} by replacing its multi-head self-attention with deeper, larger convolutional layers. 
This modification significantly increases the effective receptive fields~(ERFs), thereby matching the performance of Vision Transformers~(ViTs). 
Although ResNet~\cite{he2016deep} effectively mitigates gradient vanishing or explosion issues linked with increased depth, recent findings suggest that its receptive field does not expand significantly with deeper configurations.
Comparisons of the effective receptive fields between ResNet and RepLKNet reveal that RepLKNet exhibits a significantly larger effective receptive field than ResNet.~\cite{raghu2021vision}. 
This indicates that RepLKNet leverages large convolutional kernels to capture more comprehensive contextual information, which leads to significant improvements in feature extraction.

LSKNet~\cite{Li_2023_ICCV} marks a significant advancement in applying large convolutional kernels to object detection within remote sensing imagery, achieving notable performance improvements by capturing essential contextual details.
Additionally, Dilated Convolutions~\cite{yu2015multi} extend the receptive field without adding computational cost by incorporating gaps in the convolutional operations, broadening the contextual scope accessible to the network.
Despite their benefits, large convolutional kernels introduce increased computational demands, which can challenge real-time applications. 
Innovations such as sparse convolutions~\cite{liu2015sparse} and mixed convolutions~\cite{tan2019mixconv} have been proposed to balance computational efficiency with feature extraction capabilities.

\subsection{Multi-Kernel Convolution Mechanisms}

To address the computational overhead associated with large convolutional kernels~\cite{chen2024pelk}, the multi-kernel convolution mechanism has been proposed. 
Dynamic Convolution~\cite{chen2020dynamic} dynamically adjusts the convolutional kernel size based on the input data conditions, optimizing computational resources.
This flexibility allows the network to select the most appropriate kernel size for each specific input, thereby enhancing adaptability and performance.
Another related technique is Mixed-Dilated Convolutions(e.g., MSFF-Net~\cite{WANG2022103319}), which combines different types of dilated convolutions to improve the perception of multi-scale objects. 
By integrating various convolutional kernels, this method can extract features at multiple scales, thereby increasing the precision of object detection.
Additionally, Convolutional Kernel Fusion~\cite{huang2018convolutional} is a technique that merges different convolutional kernels into a unified operation. 
This approach reduces computational overhead while enhancing feature extraction efficiency. 
Convolutional Kernel Fusion is particularly suitable for scenarios with limited computational resources, as it maintains high efficiency while improving model performance.

\subsection{Feature Representation in Complex Backgrounds}

Enhancing feature representation within complex backgrounds is a key challenge in computer vision. 
Attention mechanisms are crucial for improving feature extraction and processing. 
Spatial attention mechanisms adaptively adjust feature map weights to focus on important areas and reduce background noise. 
For example, SE-Net~\cite{hu2018squeeze} employs a module that adjusts feature weighting based on global information, significantly enhancing performance across various tasks.
Channel attention mechanisms refine feature representation by optimizing channel weights and prioritizing crucial features for target recognition. 
CBAM~\cite{woo2018cbam} merges channel and spatial attention, greatly improving extraction accuracy and model robustness. Despite their effectiveness, integrating these mechanisms in high-resolution scenarios remains a research focus. 
ECA-Net~\cite{wang2020eca} introduces streamlined channel attention that reduces complexity while enhancing feature representation, showing excellent performance across benchmarks.

\section{Approach}

\begin{figure}[htb]
    \centering
    \includegraphics[width=1 \linewidth]{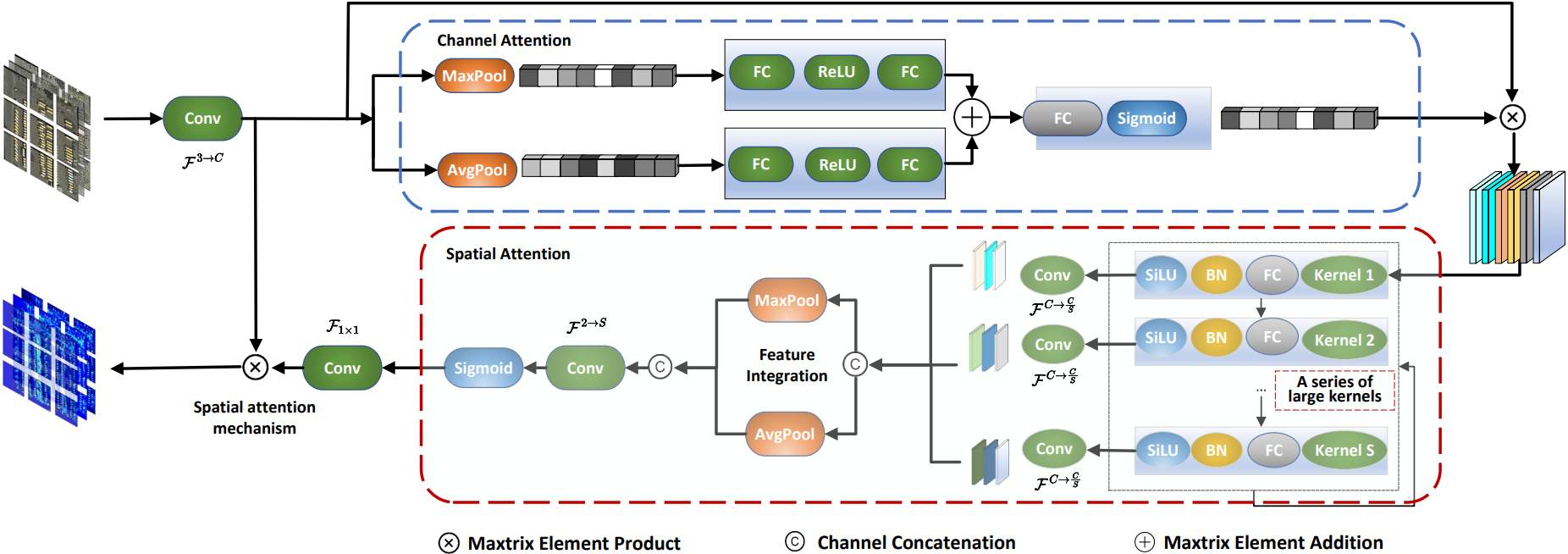}
    \caption{Overall framework of MKSNet.
    The MKSNet comprises a sequence of MKS blocks, with each block incorporating a Channel Attention module (at the top) and a Spatial Attention module (at the bottom). 
    It starts with the input image being divided into patches via a convolutional layer. 
    These patches undergo enhancement by the Channel Attention to emphasize significant channels, followed by the Spatial Attention focusing on key areas. 
    The MKSNet dynamically selects various kernel sizes to capture and integrate multi-scale contextual information, significantly improving detection performance.}
    \label{fig-framework}
\end{figure}

The MKSNet, as shown in Figure \ref{fig-framework}, consists of two primary modules: Channel Attention (CA) and Spatial Attention (SA). 
The CA module dynamically adjusts the weights of feature channels to prioritize critical information, enhancing the model's focus on relevant channel features.
Concurrently, the SA module refines the focus on key image regions, effectively reducing background interference.
Together, these modules enable MKSNet to adeptly capture and integrate multi-scale contextual information.
Finally, the feature fusion module consolidates these enriched features to boost the accuracy.

\subsection{Spatial Attention~(SA) Module}
\subsubsection{Spatial Feature Extraction}

To effectively capture multi-scale spatial information from input feature maps, we developed a comprehensive spatial convolution block that employs convolutional kernels of varying sizes and dilation rates.
We define a series of convolutional kernels, each denoted as $\mathcal F_i (X; k_i, d_i, p_i)$, with varying sizes $k_{i}$, dilation rates $d_{i}$, and a fixed stride $S$.
The size and dilation rate of these kernels increase linearly with the step-index $i$, and the total number of kernels is limited to a maximum value, $max\_size$. 
To ensure that the size of the feature map remains unchanged after convolution, we use appropriate padding $p_i$. 
The spatial convolution operation can be described as follows:
\begin{equation}
p_i = \frac {(k_i -1) \cdot d_i}{2}, \quad
k_{i} = \min(5 + 2 \cdot i, max\_size), \quad
d_{i} = i + 1
\end{equation}
where $X$ is the input feature map, and $\mathcal F_{i}$ is the feature map obtained at step $i$. 
This array of convolutional kernels effectively captures spatial information across various scales, extracting richer features critical for detecting small objects.

We propose the MKS block to extract multi-scale features from the input feature map. 
It consists of multiple convolutional layers, each with different kernel sizes and dilation rates. 
Specifically, given the input image $X\in \mathbb{R}^{C\times H\times W}$, it undergoes feature extraction through a series of spatial convolutions with varying kernel sizes. 
Each spatial convolution layer $\mathcal{F}_i^{k_i\times k_i}$ acts on $X$, followed by a $1\times 1$ convolution $\mathcal{\bm{F}}_i^{1\times 1}$  that linearly combines features across all channels. 
This process enhances non-linear transformations and promotes information fusion, culminating in a feature map $\tilde X_{i}\in\mathbb{R}^{C\times H\times W}$ for each step.
These feature maps are concatenated to form a composite feature set in $S$ batches as $\tilde X=\{\tilde{X}_1, \tilde{X}_2,\ldots,\tilde{X}_S\}$, defined as:
\begin{gather}
    \tilde X_{{i}}^{'}=BN(\mathcal F_i^{k_i \times k_i}(X))\\
    \tilde X_{{i}}=\sigma(\mathcal F_i^{1\times 1}(\tilde X_{i}^{'})), \quad i \in[1, S]
\end{gather}
where $\text{BN}(\cdot)$ denotes batch normalization, and $\sigma(\cdot)$ is the activation function.

\subsubsection{Channel Transformation}

The channel transformation module projects feature maps extracted at different scales to a uniform channel dimension, facilitating processing by the attention mechanism.
Each scale's feature map transforms a $1\times 1$ convolutional layer $\mathcal{F}_i^{1 \times 1}$, \textit{i.e.}, $\mathcal{T}_i= \mathcal{F}_i^{1\times 1} (\tilde X_i), \mathcal{T}_i  \in \mathbb{R}^{\frac{C}{S}\times H \times W}$, facilitating subsequent processing by the attention mechanism. 
Then, we concatenated these transformed feature maps along the channel dimension as $\mathcal{T} = \text{\textit{Concat}}(\mathcal{T}_1,\mathcal{T}_2,\ldots,\mathcal{T}_S ), \mathcal{T}\in \mathbb{R}^{C\times H\times W}$

\subsubsection{Spatial Attention Mechanism}

The SA module employs a spatial attention mechanism to enhance feature map representation.
It calculates the average and maximum values across all transformed feature maps, generating an attention weight map through a convolutional layer $\mathcal F^{2\rightarrow S}$ and a sigmoid activation function. 
The specific steps are as follows:
\begin{gather}
    \mathcal{M} = Concat(Mean(\mathcal T ), Max(\mathcal T)), \quad \mathcal M \in \mathbb{R}^{2 \times H  \times W}\\
    \mathcal Sig = Sigmoid(\mathcal F^{2 \rightarrow S}(\mathcal M)), \quad \mathcal Sig \in \mathbb{R}^{S\times H\times W} \label{eq:sig}
\end{gather}
where $ \mathcal Sig $ is the attention weight map, $Mean(\cdot)$ represents average pooling along the channel dimension, $Max(\cdot)$ denotes maximum pooling along the channel dimension, and $Concat(\cdot)$ is used for concatenation along the channel dimension.
This attention mechanism focuses the network on critical regions by weighting feature maps, enabling MKSNet to effectively distinguish small objects from complex backgrounds.

The final weighted feature map $\mathcal Sig$ is obtained by element-wise multiplication of the attention weight map with the feature maps, summing them to determine their contribution to the final output. 
The processed feature map is then further refined using a convolutional operation, integrating with the input feature map to produce the output:
\begin{gather}
    \mathcal{P} = \sum_{i=1}^{S}\mathcal T_i \odot \mathcal Sig_i \\
    \mathcal O = X \odot \mathcal F^{1 \times 1} (\mathcal P )
\end{gather}
where $ \mathcal{P}$ is the weighted attention feature map, $ \mathcal Sig_i$ denotes the attention weight for the feature map at the $i$-th scale, $\odot$ represents element-wise multiplication, and $\mathcal{O}$ is the final output result.

\subsection{Channel Attention~(CA) Module}

To enhance the model's precision in emphasizing vital features, we introduce the channel attention mechanism, inspired by the principles of Squeeze-and-Excitation Networks (SENet)~\cite{hu2018squeeze}.
This mechanism dynamically adjusts the channel strengths of feature maps by assigning learned weights to each channel, which amplifies critical information within the overall feature representation.
By integrating global pooling and fully connected layers, it generates a specific weight coefficient for each channel, thereby fine-tuning the feature maps to highlight essential details and boost the model's expressiveness and accuracy.

\subsubsection{Channel Feature Extraction}

Global pooling operations compress the spatial dimensions of the input feature map to $1\times 1$, effectively summarizing global statistics for each channel. 
Global average pooling calculates the mean value across each channel, while global max pooling determines the maximum value, collectively providing a comprehensive snapshot of global channel-level information. 
Specifically, consider $X \in \mathbb{R}^{B \times C \times H \times W}$ as the input feature map, where $B$ is the batch size, $C$ represents the number of channels, and $H$ and $W$ are the height and width of the feature map, respectively. 
Applying global average and max pooling to $X$ reduces its spatial dimensions from $H \times W$ to $1 \times 1$, as outlined below:
\begin{gather}
    A = AvgPool(X), \quad A \in \mathbb{R}^{B \times C}\\
    M = MaxPool(X), \quad M \in \mathbb{R}^{B \times C}
\end{gather}
where $A$ is the channel feature vector obtained through average pooling, $M$ is the channel feature vector obtained through max pooling, $AvgPool(\cdot)$ denotes the average pooling function along the channel dimension, and $MaxPool(\cdot)$ denotes the max pooling function along the channel dimension.

\subsubsection{Channel Feature Transformation}

The global features $A$ and $M$ are first compressed via two separate fully connected layers, reducing the channel dimension from $C$ to $\frac{C}{r}$ by a downscaling factor of $r$. 
After activation, they are expanded back to the original dimension and averaged to form a unified channel feature. Subsequently, the two feature representations are combined using weighted averaging to obtain the fused feature. 
Finally, a fully connected layer is employed to adjust the feature dimension to $ C\times 1 \times 1 $. 
The specific operations are as follows:
\begin{gather}
    \tilde{A} = \sigma(FC_1(A)), \quad \tilde{A} \in \mathbb{R}^{B \times C} \\
    \tilde{M} = \sigma(FC_2(M)), \quad \tilde{M} \in \mathbb{R}^{B \times C}\\
    \tilde O=FC_3(\frac{\tilde{A}+\tilde{M}}{2}), \quad \tilde O \in \mathbb{R}^{B\times C \times 1 \times 1}
\end{gather}
where $FC_1(\cdot)$ and $ FC_2(\cdot)$ are two distinct fully connected layers that reduce the number of channels from $C$ to $\frac{C}{r}$, with $r$ being the reduction factor. 
$\sigma(\cdot)$ denotes the activation function. 
$\tilde A$  and $\tilde M$ are the compressed global average and max feature vectors, respectively.
$FC_3(\cdot)$ is a fully connected layer that restores the channel dimension to $C\times 1\times 1$, producing $\tilde O$, the refined global channel feature vector after transformation.

\subsubsection{Channel-wise weighting}
The refined channel features $\tilde{O}$ are then applied to the input feature map $X$ through element-wise multiplication, adjusting each channel's intensity based on its relevance:
\begin{gather}
    O = X \odot Sigmoid(\tilde{O}), \quad O \in\mathbb{R}^{ B\times C\times H\times W}
\end{gather}
where $\odot$ denotes the element-wise multiplication operation; 
the sigmoid function maps the attention weight values to the range $[0, 1]$, allowing smooth adjustment of the strength of each channel; 
and $ O $ represents the final feature map after channel-wise weighting.

\section{Experiments}
\subsection{Dataset and Evaluation Metrics}
\subsubsection{Datasets}\label{section:datasets}

DOTA-v1.0~\cite{Xia_2018_CVPR} is a comprehensive dataset for object detection in remote sensing imagery, including Helicopter (HC), Storage Tank (ST), Tennis Court (TC), Plane (PL), Large Vehicle (LV), Baseball Diamond (BD), Swimming Pool (SP), Ground Track Field (GTF), Bridge (BR), Roundabout (RA), Ship (SH), Soccer-ball Field (SBF), Basketball Court (BC), Small Vehicle (SV), and Harbor (HA). 
HRSC2016~\cite{liu2016ship} is a high-resolution remote sensing dataset for ship detection, containing 1061 images from major global ports and 2976 annotated ship instances, designed for detailed detection and classification tasks.

\subsubsection{Evaluation metrics}

In this study, we use Mean Average Precision (mAP) to evaluate network performance. 
mAP is a pivotal metric in object detection evaluations.
It gauges a model's comprehensive performance by averaging the Average Precision (AP) scores across multiple categories~\cite{lin2014microsoft}.
Average Precision (AP) is calculated by constructing the Precision-Recall (PR) curve and computing the area beneath this curve~\cite{everingham2009pascal}. 
The AP for each category is derived using the integral $AP=\int_{0}^{1} Prc(r)\, dr$, where $Prc(r)$ indicates the precision at a given recall level $r$.
To determine mAP, AP is first computed for each category. 
Subsequently, these AP values are aggregated to provide an average $mAP=\frac{1}{C}\sum_{i=1}^{C}AP_i$, where $C$ is the total number of categories, and $AP_i$ is the AP for the $i_{th}$ category.

\subsection{Implementation Details}

The MKSNet was rigorously evaluated using the datasets mentioned above, which were divided into training, validation, and testing sets to ensure a comprehensive performance evaluation and robust model generalization.
Our model was implemented using the PyTorch framework based on the Ubuntu 22.04 system and trained with the AdamW optimizer~\cite{kingma2014adam}. 
We set the initial learning rate to 0.0004, the momentum coefficients to (0.9, 0.999), and the weight decay to 0.05. 
Training, validation, and testing of the network were conducted within the Oriented RCNN framework~\cite{hu2018relation}, which effectively handles target orientation by integrating direction-aware convolutional networks and regression models, thereby enhancing detection accuracy.

Training was conducted on three NVIDIA GeForce RTX 4090 GPUs with a batch size of two per GPU, over 300 epochs for each dataset. 
Given the high-resolution characteristics of the DOTA-v1.0 dataset, a specialized preprocessing technique was utilized. 
This technique involved segmenting large images into smaller, overlapping units to ensure no critical information was lost at the edges, enhancing the detection efficacy in boundary regions and maintaining high fidelity in feature representation.

\subsection{Comparison with State-of-the-Arts}

We compare MKSNet with other state-of-the-art methods on the DOTA-v1.0 and HRSC2016 datasets, demonstrating its superior performance.
Table~\ref{tab:dota-sota} and Table~\ref{tab:hrsc-sota} detail our findings, showcasing MKSNet's new state-of-the-art achievements in object detection within remote sensing imagery.
\begin{table}[h]
    \renewcommand{\arraystretch}{2.5}
    \setlength{\tabcolsep}{4pt}    
 \fontsize{30}{20}\selectfont
\resizebox{\linewidth}{!}{
\begin{tabular}{l|c|c|c|c|c|c|c|c|c|c|c|c|c|c|c|c|c|c}
\hline
\textbf{Method} & \textbf{mAP}  & \textbf{Params}& \textbf{FLOPs}  & \textbf{PL} & \textbf{BD} & \textbf{BR} & \textbf{GTF} & \textbf{SV} & \textbf{LV} & \textbf{SH} & \textbf{TC} & \textbf{BC} & \textbf{ST} & \textbf{SBF} & \textbf{RA} & \textbf{HA} & \textbf{SP} & \textbf{HC} \\
\Xhline{2px}
\textbf{RoI Trans.~\cite{ding2019learning}} & 75.79 & 57.4M& \underline{200G}& 86.85 & 83.36 & 52.63 & \textbf{80.61} & 79.43 & 78.47 & 76.67 & 85.90 & 84.72 & \underline{83.31} & 60.27 & 64.27 & 70.21 & 60.13 & 72.58 \\
\textbf{DAFNe~\cite{lang2021dafne}} & 76.72 & 70.7M & 392G & \textbf{88.12} & 84.58 & 52.80 & 80.38 & \textbf{82.46} & 80.59 & 84.29 & 86.01 & 80.86 & 79.95 & 58.44 & \textbf{68.30} & 74.14 & \textbf{79.77} & \textbf{75.47} \\
\textbf{CenterMap~\cite{wang2020learning}} & 73.59 & \underline{42.7M}& 204G& 84.15 & 85.13 & 52.91 & 76.22 & 71.18 & 77.50 & \underline{86.12} & 84.98 & \underline{87.24} & 72.36 & 63.31 & 60.23 & 64.28 & 70.01 & 71.85 \\
\textbf{R3Det~\cite{lyu2022rtmdet}} & 73.10 & 43.1M& 328G& \underline{87.48} & 82.42 & 46.31 & 64.87 & 77.15 & \underline{81.92} & 85.88 & \textbf{89.15} & 83.88 & \textbf{84.30} & 63.70 & 61.30 & \underline{65.80} & 76.70 & 70.72 \\
\textbf{O-RCNN~\cite{xie2021oriented}} & \underline{76.12} & 44.8M& 203G& 86.71 & \textbf{86.62} & \underline{56.27} & 77.69 & 75.67 & 80.00 & 82.46 & 83.70 & 86.65 & 82.12 & \underline{64.33} & \underline{67.88} & 75.52 & 72.21 & 72.37 \\
\textbf{MKSNet} (\textbf{ours})& \textbf{78.77} & \textbf{40.7M}& \textbf{181G}& 85.66 & \underline{85.43} & \textbf{56.89} & \underline{80.43} & \underline{80.51} & \textbf{82.90} & \textbf{86.35} & \underline{88.11} & \textbf{87.32} & 81.09 & \textbf{69.92} & 66.94 & \textbf{76.32} & \underline{74.11} & \underline{74.23} \\
\hline
\end{tabular}
}
\caption{
Category-wise and overall mAP comparison for object detection on the DOTA-v1.0 dataset. 
Bold indicates the best performance, while underlined numbers indicate the second-best.}
\label{tab:dota-sota}
\vspace{-1em}
\end{table}
\subsubsection{Results on DOTA-v1.0} 
The MKSNet, with its novel backbone network, showed significant improvements over previous state-of-the-art methods. 
It achieved a $2.6\%$ increase in mAP compared to models using a standard ResNet-50 backbone, showcasing faster convergence and robust performance even with fewer training iterations. 
This advantage is particularly beneficial in scenarios where computational resources and time are limited.
Additionally, our method outperforms previous SOTA models in identifying more challenging targets, such as bridges, small vehicles, basketball courts, and roundabouts.
\begin{table}[h]
\renewcommand{\arraystretch}{1}
\setlength{\tabcolsep}{25pt}    
\resizebox{\linewidth}{!}{
\begin{tabular}{l|c|c|c|c|}
\hline
\textbf{Method} & Params& FLOPs& mAP(150)& mAP(300)\\
\Xhline{1px}
\textbf{RoI Trans.~\cite{ding2019learning}}& 57.4M& \underline{200G}& 66.87& 76.25\\
\textbf{ReDet~\cite{han2021redet}}& \textbf{35.9M}& 217G& 67.79& 79.42\\
\textbf{CenterMap~\cite{wang2020learning}} & 42.7M& 204G& 66.12& 76.87\\
\textbf{R3Det~\cite{lyu2022rtmdet}} & 43.1M& 328G& 69.60& 78.56\\
\textbf{O-RCNN~\cite{xie2021oriented}} & 44.8M& 203G& \underline{71.33}& \underline{83.89}\\
\textbf{MKSNet} (\textbf{ours})& \underline{40.7M}& \textbf{181G}& \textbf{71.95}& \textbf{84.31}\\
\hline
\end{tabular}
}
\caption{Overall mAP comparison for object detection on the HRSC2016 dataset. 
Bold indicates the best performance, and underlined figures denote the second-best.}
\label{tab:hrsc-sota}
\end{table}
\subsubsection{Results on HRSC2016.}
Employing our customized MKSNet backbone, the model reached mAPs of $71.95\%$ at 150 epochs and $84.31\%$ at 300 epochs, surpassing other state-of-the-art models utilizing standard backbones.
This marks an improvement of $0.6\%$ and $1.58\%$ over other state-of-the-art models using ResNet-50.
Our approach consistently outperforms competing models in handling multi-scale objects across various datasets, demonstrating enhanced stability and superior performance in high-resolution remote sensing imagery.
Remote sensing datasets typically feature high-resolution images with small targets against complex backgrounds, challenging conventional object detection methods. 
While previous approaches have shown decent performance, they often struggle with slow convergence on multi-scale objects. 
In contrast, our method not only outperforms many state-of-the-art techniques in detection accuracy but also achieves faster convergence, delivering robust results more efficiently.

\section{Ablation Study}

To systematically evaluate the contributions of the Spatial Attention Module (SA) and Channel Attention Module (CA) within our proposed MKSNet framework, we conducted an extensive ablation study. 
The ablation experiments are designed to isolate the effects of each module and evaluate their impact on the overall detection performance. 

We evaluated four configurations: the Base Model (baseline without SA and CA, using ResNet-50 for feature extraction); 
Base + SA (incorporating the Spatial Attention mechanism to assess the impact of large kernels on small object detection); 
Base + CA (integrating the Channel Attention mechanism to evaluate the effect of channel-wise attention on feature representation); 
and Base + SA + CA (the full model combining both mechanisms to examine their synergistic effect on detection performance). 
Each configuration was trained for 100 epochs, and the results are summarized in Table~\ref{tab:ablation}.
\vspace{-1em} 
\begin{table}
 %    \renewcommand{\arraystretch}{2.5}
 %    \setlength{\tabcolsep}{2pt}    
 % \fontsize{6}{5}\selectfont
    \centering
    \begin{tabular}{ccc|ccc}
    \Xhline{1px}
        Base&  SA &  CA&  mAPs(\%)& $\Delta$ \\
    \hline
        \checkmark &  &  & 62.7 & 0.0\\
        \checkmark & \checkmark &  & 66.4 & +3.7 \\
        \checkmark &  & \checkmark & 64.3 &  +1.6 \\
        \checkmark & \checkmark & \checkmark & 69.1 & +6.4\\
    \Xhline{1px}
    \end{tabular}
    % \vspace{1em} 
    \caption{Ablation studies of MKSNet with different component integrations.}
    \label{tab:ablation}
\end{table}

\begin{figure}[ht]
    \centering
    \includegraphics[width=0.6 \linewidth]{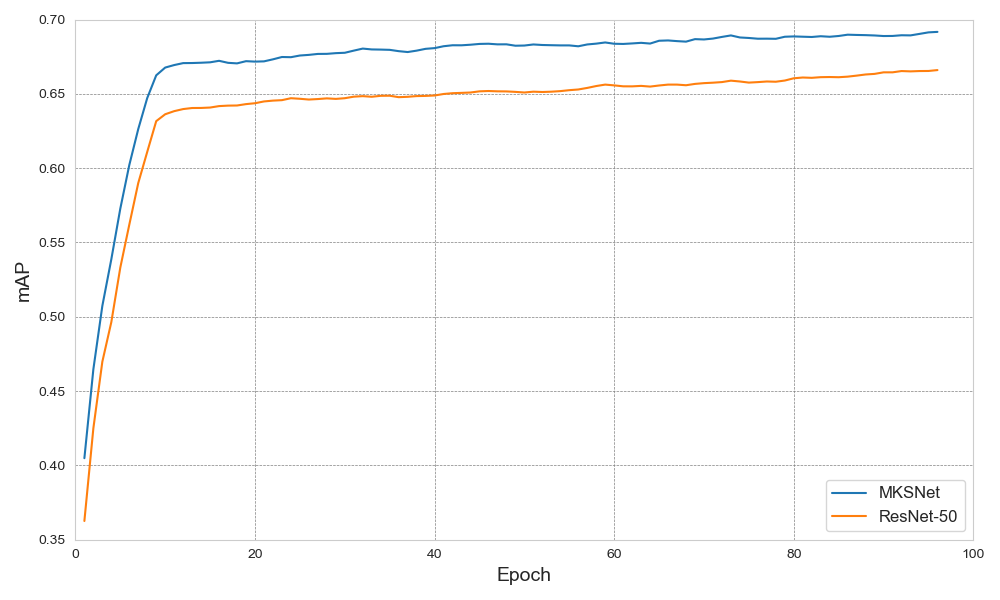}
    \caption{mAP comparison over 100 epochs between MKSNet and ResNet-50.}
    \label{fig-curve}
\end{figure}

Our model configuration significantly outperforms the ResNet-50 baseline in terms of mAP.
Both setups show a rapid increase in mAP within the initial 20 epochs; 
however, MKSNet continues to improve, reaching a higher stable mAP by the 100th epoch, while ResNet-50's performance plateaus at a lower level.
This demonstrates that MKSNet more effectively captures complex contextual information and optimizes feature representation, resulting in enhanced overall detection performance. 
The comparative mAP progressions over 100 epochs are illustrated in Figure~\ref{fig-curve}.

\section{Conclusions}

In this paper, we present MKSNet, a novel neural network architecture aimed at improving small object detection in remote sensing imagery. Traditional convolutional networks often struggle with information loss and background noise when detecting small objects. To overcome these challenges, we introduce the Multi-Kernel Selection (MKS) mechanism, which utilizes multi-scale large kernels to capture spatial details more effectively, significantly enhancing detection accuracy. Additionally, a channel attention mechanism dynamically adjusts feature channel weights, further optimizing feature extraction and improving both accuracy and robustness. Extensive evaluations on the DOTA-v1.0 and HRSC2016 datasets demonstrate that MKSNet outperforms current state-of-the-art methods in small object detection, offering valuable insights and a robust framework for future research in remote sensing image analysis.\\

\noindent
\textbf{Acknowledgment}
This work was supported by the Industry-University-Institute Cooperation Foundation of the Eighth Research Institute of China Aerospace Science and Technology Corporation (No. SAST2022-049), and the National Natural Science Foundation of China under No. 62472033 and No. 61972036.

\newpage

\bibliographystyle{splncs04} 
\bibliography{ref.bib}

\end{document}